\documentclass[10pt,journal,compsoc]{IEEEtran}
\usepackage{amsmath,amssymb,amsthm}
\usepackage{booktabs}
\usepackage{hyperref}
\usepackage{cite}
\usepackage{graphicx}
\usepackage{xcolor}
\usepackage{tikz}
\usetikzlibrary{positioning,arrows.meta,decorations.pathreplacing,calc,fit,backgrounds}

\newtheorem{theorem}{Theorem}[section]
\newtheorem{definition}[theorem]{Definition}
\newtheorem{proposition}[theorem]{Proposition}

\theoremstyle{remark}

\begin{document}

\title{Structural Decoupling: A Scaffold-Flow Theory of Generalization and Alignment}

\author{%
\IEEEauthorblockN{Xin Li}\\
\IEEEauthorblockA{Department of Computer Science, \\University at Albany, NY 12222}
\thanks{This work is partially supported by NSF awards IIS-2401748 and BCS-2401398. The author has used ChatGPT 5.5 and Claude 4.7 models to assist in the development of theoretical ideas, mathematical proofs in the Appendix, and visual illustrations presented in this paper.}
}

\IEEEtitleabstractindextext{%
\begin{abstract}
Learning in non-stationary and multi-context environments requires more than ordinary within-task generalization. A system must also discover which contexts exist, route inputs to the correct context, preserve old contexts, and revise the context library when the environment changes. This paper presents Structural Learning Theory (StrLT) as a framework of filling this missing structural gap. StrLT complements Vapnik's Statistical Learning Theory (SLT): SLT governs the \emph{funnel}, prediction or control within a fixed regime; while StrLT governs the \emph{trap}, the discovery and maintenance of structural regimes.
The core StrLT object is \emph{width}, the minimum number of locally feasible contexts needed to cover a problem. We summarize three basic results: width is incomparable with VC dimension; learning exhibits a phase transition at the true width; and width can be estimated by a contractive-similarity (CS) operator that converts task-induced non-contractivity into spectral separation. Under the StrLT framework, we explain how fixed-class structural learnability leads to a \emph{structural decoupling principle}: the mechanisms that maintain the structural scaffold should not be trained by the same gradients that optimize within-context flow. This principle motivates a scaffold-flow model in which alignment and generalization separate architecturally. Finally, we argue that several safety failures, including hallucination, reward-model boundary errors, and deceptive alignment, can be interpreted as scaffold-resolution or scaffold-preservation failures rather than merely output-level prediction errors.
\end{abstract}

}

\maketitle
\IEEEdisplaynontitleabstractindextext

\begin{figure*}[h]
\centering
\resizebox{\linewidth}{!}{
\begin{tikzpicture}[
    font=\small,
    node distance=1.4cm,
    >=stealth,
    every node/.style={align=center},
    box/.style={
        draw,
        rounded corners,
        thick,
        minimum width=2.8cm,
        minimum height=1.0cm,
        fill=gray!8
    },
    trapbox/.style={
        draw,
        rounded corners,
        thick,
        minimum width=3.1cm,
        minimum height=1.2cm,
        fill=orange!12
    },
    funnelbox/.style={
        draw,
        rounded corners,
        thick,
        minimum width=3.1cm,
        minimum height=1.2cm,
        fill=blue!10
    },
    smallbox/.style={
        draw,
        rounded corners,
        thick,
        minimum width=2.2cm,
        minimum height=0.85cm,
        fill=white
    },
    arrow/.style={->, thick},
    dashedarrow/.style={->, thick, dashed}
]

\node[box] (input) at (0,0) {
    Input \(x\in X\)\\
    structured problem
};

\node[trapbox, right=1.9cm of input] (trap) {
    \textbf{Trap}\\
    identify context\\
    \(c=\Sigma(x)\)
};

\node[smallbox, above right=0.7cm and 1.5cm of trap] (c1) {
    Cell \(U_1\)
};
\node[smallbox, right=1.5cm of trap] (c2) {
    Cell \(U_2\)
};
\node[smallbox, below right=0.7cm and 1.5cm of trap] (cw) {
    Cell \(U_w\)
};

\node[funnelbox, right=1.7cm of c1] (f1) {
    \textbf{Funnel \(1\)}\\
    local contraction\\
    \(G_1\)
};
\node[funnelbox, right=1.7cm of c2] (f2) {
    \textbf{Funnel \(2\)}\\
    local contraction\\
    \(G_2\)
};
\node[funnelbox, right=1.7cm of cw] (fw) {
    \textbf{Funnel \(w\)}\\
    local contraction\\
    \(G_w\)
};

\node[box, right=1.7cm of f2] (output) {
    Prediction / action\\
    \(\hat y = G_{\Sigma(x)}(x)\)
};

\draw[arrow] (input) -- (trap);
\draw[arrow] (trap) -- (c1);
\draw[arrow] (trap) -- (c2);
\draw[arrow] (trap) -- (cw);

\draw[arrow] (c1) -- (f1);
\draw[arrow] (c2) -- (f2);
\draw[arrow] (cw) -- (fw);

\draw[arrow] (f1.east) -- ++(0.7,0) |- (output.north west);
\draw[arrow] (f2) -- (output);
\draw[arrow] (fw.east) -- ++(0.7,0) |- (output.south west);

\node[above=0.2cm of trap, text=orange!70!black] {
    structural / topological axis
};

\node[above=0.2cm of f1, text=blue!70!black] {
    metric / statistical axis
};

\draw[decorate, decoration={brace, amplitude=6pt}, thick]
    ($(c1.north west)+(-0.2,0.35)$) --
    ($(cw.south west)+(-0.2,-0.35)$)
    node[midway, left=0.35cm] {
        width\\
        \(w(P)\)
    };


\end{tikzpicture}
}
\caption{
Trap-funnel decomposition in StrLT. The trap is the structural problem of identifying the correct basin or context \(c=\Sigma(x)\). Once the context is known, the funnel is the metric problem of learning or applying a local contraction \(G_c\). Note that there is no task-loss gradient into the scaffold due to structural decoupling by architecture.
}
\label{fig:trap-funnel}
\end{figure*}

\section{Introduction}
\label{sec:intro}

\IEEEPARstart{I}{ntelligent} systems increasingly operate in environments that are structured, non-stationary, and multi-regime \cite{sugiyama2012machine}. An autonomous vehicle moves among highway driving, urban navigation, parking, and emergency scenarios \cite{yurtsever2020survey}. A medical AI serves patient populations with different latent mechanisms \cite{tjoa2020survey}. A language model shifts among mathematical reasoning, factual recall, instruction following, and tool use \cite{xia2026skillrl}. In all of these cases, the system faces two problems at once: it must determine \emph{which regime is active} and then plan or act well \emph{within that regime}. These two problems are not the same. The first is \emph{structural}: how many regimes exist, how should they be represented, and when should routing change? The second is \emph{metric}: once a regime is fixed, how hard is optimization or generalization inside it?

Classical Statistical Learning Theory (SLT) by Vapnik \cite{vapnik1998statistical} gives a powerful theory of the metric problem. VC dimension, Rademacher complexity, covering numbers, and PAC learnability characterize generalization within a fixed hypothesis class and a fixed data-generating regime \cite{vapnik1998statistical,vapnik1971uniform,shalev2014understanding,bartlett2002rademacher}. Control theory similarly analyzes regulation and optimization once the relevant state variables and regime are known \cite{kalman1960new,bertsekas2012dynamic}. What these theories do not directly characterize is the structural problem that comes before local optimization: \emph{discovering the contexts} in which local learning is valid and preserving those contexts under non-stationarity.

The problem of structural discovery appears in several traditions under different names. Cybernetics emphasizes feedback, mismatch, and reconfiguration \cite{ashby1956introduction,ashby1952design}. Complementary Learning Systems (CLS) separates fast hippocampal indexing from slow cortical consolidation \cite{mcclelland_why_1995,oreilly2014complementary}. Mixture-of-experts (MoE) \cite{jacobs1991adaptive,li2025theory} architectures introduce routers and specialized experts, but are typically analyzed as structured hypothesis classes within an SLT-compatible framework. Continual learning exposes the failure mode most sharply \cite{wang_comprehensive_2024}: without stable context discovery and preservation, new learning interferes with old learning and catastrophic forgetting occurs \cite{kirkpatrick_overcoming_2017}. The central motivation of this paper is that these perspectives are converging on the same missing axis of learning structural complexity.

Structural Learning Theory (StrLT) fills this missing gap with geometric intuition of the \emph{trap-funnel} decomposition (Fig. \ref{fig:trap-funnel}). The \emph{trap} is the structural problem of identifying the correct context, basin, or scaffold; the \emph{funnel} is the metric problem of optimizing prediction or control after that context is known. The central complexity parameter is \emph{width}: the minimum number of locally feasible contexts required to cover a problem. Width is not a substitute for VC dimension but an orthogonal complexity measure: a problem can have a large width with a small VC dimension, or a large VC dimension with a unit width. The VC-width separation has profound implications for continual or lifelong learning \cite{wang_comprehensive_2024}. The fundamental tradeoff between stability and plasticity \cite{mermillod2013stability} is resolved by complementary learning systems \cite{mcclelland_why_1995}, admitting a StrLT interpretation from the \emph{structural decoupling} perspective, which sets up the stage for this paper. 

The structural-metric separation changes how we should think about architecture in both artificial and natural intelligent systems \cite{minsky_steps_1961}. If the trap and the funnel correspond to different sources of error, then a single entangled mechanism should not be expected to solve both reliably. Along this line of reasoning, StrLT leads naturally to a \emph{scaffold-flow} view \cite{li_two_2026}: the scaffold maintains the discrete structural organization of contexts, while the flow performs continuous prediction, optimization, or generation within a selected context. Generalization depends on the flow being statistically well behaved, but it also depends on the scaffold being correctly resolved and stable \cite{hendrycks_many_2021}.
If the scaffold drifts, collapses, or routes an input to the wrong context, then the model may generate a locally plausible output under an inappropriate structural interpretation \cite{gama2004learning}. 
Such errors are not prediction errors along the metric axis but failures of
contextualization along the structural axis, which become directly relevant to
alignment. If robust generalization requires a structurally decoupled scaffold,
then alignment also requires preserving and auditing that scaffold. A model may
produce locally plausible outputs while relying on an unstable or misresolved
context structure, analogous to goal misgeneralization and inner-alignment
failures in which capabilities generalize while the intended objective or
context does not~\cite{langosco2022goal,shah2022goal,hubinger2019risks,hendrycks2021unsolved}.

StrLT provides a conceptual framework for studying the structural failure mode: flow alignment controls behavior inside a context, while scaffold alignment controls the contexts in which behavior is interpreted, routed, and generalized.
This paper makes four technical contributions. 
\begin{itemize}
    \item We formalize the structural axis of learning through \emph{width}, defined as the minimum number of jointly contractive and low-risk cells needed to cover a problem, and proves that width is incomparable with VC dimension: structural complexity can diverge while classical capacity remains bounded, and conversely.
    \item We develop an empirical theory of width estimation through the contractive-similarity (CS) operator, whose Laplacian converts prediction incompatibility into spectral separation; this yields split--merge criteria, uniform width-estimation guarantees, and structural ERM consistency under explicit regularity assumptions.
    \item We introduce the metric slingshot, a funnel-side construction showing how low-dimensional latent contraction maps can transfer contraction and risk guarantees back to the original problem through a controlled-distortion embedding.
    \item We derive a structural decoupling principle from the fixed-class structural fundamental theorem: structural assignment complexity and within-cell predictor complexity contribute additively to generalization, motivating a scaffold--flow architecture in which alignment-sensitive context structure is separated from metric prediction flow. 
\end{itemize}
Together, these results turn the trap-funnel decomposition from an analogy into a mathematical pipeline: identify width, estimate and refine structural basins, solve local funnels, and separate alignment from generalization by architecture.

The paper proceeds as follows. Section~\ref{sec:traditions} positions continual learning as the convergence point of cybernetics, SLT, CLS, and modular architectures. Section~\ref{sec:strlt-core} summarizes the core results of StrLT: width, VC--width separation, phase transition at the true width, CS-based width estimation, and the metric slingshot for reducing funnel cost. Section~\ref{sec:structural-fundamental-theorem} connects these results to fixed-class structural learnability and derives the structural decoupling principle. Section~\ref{sec:scaffold-flow} turns that principle into a scaffold--flow model: the scaffold maintains context, routing, and boundary structure, while the flow performs within-scaffold statistical generalization. Section~\ref{sec:alignment-scaffold} applies this distinction to AI safety, arguing that alignment failures should be analyzed not only as output-level flow errors but also as scaffold-resolution, scaffold-boundary, and scaffold-preservation failures. Section~\ref{sec:future} outlines research directions, and Section~\ref{sec:conclusion} concludes.

\section{Continual Learning as the Convergence}
\label{sec:traditions}

The need for StrLT is clearest in continual learning. A learner receiving a non-stationary stream must solve more than a sequence of supervised prediction problems. It must decide whether the current data belong to an existing context, whether a new context should be created, whether old contexts should be preserved or merged, and how local predictors can be updated without corrupting previously learned structure. In the language of this paper, continual learning exposes the missing \emph{structural} axis: before the system can optimize within a regime, it must discover which regime it is in.
This challenge brings together three older traditions. Cybernetics studies adaptive regulation under feedback and already distinguishes fast regulation from slower reconfiguration \cite{ashby1952design,ashby1956introduction}. Statistical Learning Theory (SLT) gives the mathematical theory of risk, capacity, and generalization once the hypothesis class and data-generating regime are fixed \cite{vapnik1998statistical,devroye2013probabilistic}. Complementary Learning Systems (CLS) provides a biological architecture in which fast hippocampal indexing coexists with slow cortical consolidation \cite{mcclelland_why_1995,oreilly2014complementary}. Mixture-of-experts (MoE) models \cite{jacobs1991adaptive} can be viewed as an SLT-compatible architectural mechanism in which the experts and router pattern define a structured hypothesis class. 
Table~\ref{tab:traditions} summarizes the convergent positioning. The central claim is that continual learning requires the convergence of cybernetics (feedback and stability), SLT (finite-sample generalization), and CLS (fast-slow architectural separation).

\begin{table*}[t]
\centering
\caption{Continual learning as a convergence point for cybernetics, SLT (MoE is treated as an SLT-compatible architectural template), and CLS. }
\label{tab:traditions}
\small
\begin{tabular}{p{2.55cm}p{2.45cm}p{3.85cm}p{3.85cm}p{2.65cm}}
\toprule
\textbf{Tradition} & \textbf{Central object} & \textbf{Contribution to continual learning} & \textbf{Limitation exposed by non-stationarity} & \textbf{StrLT role} \\
\midrule
Cybernetics & Feedback, requisite variety, ultrastability & Models adaptation as regulation plus reconfiguration; motivates fast inner loops and slower structural loops & Lacks statistical sample complexity, generalization bounds, and a canonical measure of structural complexity & Supplies the control-theoretic intuition for trap detection and structural feedback \\
\addlinespace
Statistical Learning Theory & Risk, capacity, VC dimension, PAC learnability & Characterizes finite-sample generalization within a fixed regime or hypothesis class & Assumes the structural regime is fixed; does not count or discover contexts & Governs the \emph{funnel}: learning inside each discovered cell \\
\addlinespace
Complementary Learning Systems & Hippocampal indexing and cortical consolidation & Provides a biological fast--slow architecture for context separation, replay, and gradual metric learning & Lacks a formal complexity measure and learnability theorem for structural discovery & Motivates structural decoupling between trap indexing and funnel consolidation \\
\bottomrule
\end{tabular}
\end{table*}

\noindent\textbf{Cybernetics: Feedback and Structural Reconfiguration}
Ashby's Law of Requisite Variety \cite{ashby1956introduction} states that a regulator controlling an environment with variety must possess sufficient internal variety to respond. This is an early statement of the structural problem: a system whose internal repertoire is too small cannot regulate an environment with too many qualitatively distinct regimes. Ashby's ultrastability \cite{ashby1952design} further separates adaptation into two loops: a fast loop that adjusts continuous variables within the current configuration, and a slower loop that changes the configuration when the fast loop fails.
In StrLT terms, the fast loop is a funnel mechanism and the slow loop is a trap mechanism. The former optimizes within a local basin; the latter detects when the current basin assignment is wrong and changes the system's structural state. Cybernetics therefore provides the architectural blueprint for StrLT, but not the learning theory. It does not specify a sample complexity for discovering regimes, a phase transition at the correct number of regimes, or a statistical criterion for when structural reconfiguration is possible.

\noindent\textbf{Statistical Learning Theory: The Funnel Theory}
Vapnik and Chervonenkis \cite{vapnik1971uniform} established the foundations of statistical learning through uniform convergence, VC dimension, and risk minimization; later treatments connect these ideas to PAC learnability and Rademacher complexity \cite{vapnik1998statistical,shalev2014understanding}. These results give a complete theory for the metric axis: if the data distribution, loss, and hypothesis class are fixed, they characterize how many samples are needed to learn within that class.
This strength is also the boundary of SLT. It assumes that the learner already knows the regime in which the samples should be interpreted. Vapnik's structural risk minimization is a capacity-control principle over nested hypothesis classes; it is not a theory of discovering the structural partition of a non-stationary problem. Thus SLT governs the \emph{funnel}: once the correct local context is known, it tells us how hard prediction is inside that context. It does not answer the trap question: how many contexts exist, and how can the learner identify them from data?

\noindent\textbf{Mixture-of-experts models: Modular Architecture} Mixture-of-experts models \cite{jacobs1991adaptive,shazeer2017outrageously,fedus2022switch} provide a practical modular architecture: a router or gating function assigns inputs to specialized experts. This resembles the StrLT decomposition because the router plays a trap-like role and each expert plays a funnel-like role. However, MoE is usually treated as a structured model family within the broader SLT-compatible landscape. The router, sparsity pattern, and expert class define a hypothesis class, and the main questions are capacity, optimization, load balancing, and statistical efficiency.
From the StrLT viewpoint, MoE is an implementation template, not the foundational theory itself. Standard MoE models do not specify the intrinsic width of the problem, do not prove that performance changes sharply at a critical number of experts, and often train routers and experts jointly through shared task-loss gradients. This coupling can blur the distinction between structural discovery and within-expert learning. StrLT supplies the missing criterion: routing is structurally adequate only when the allocated contexts match or exceed the problem width and when the trap and funnel are sufficiently decoupled.

\noindent\textbf{Complementary Learning Systems: The Biological Separation}
Complementary Learning Systems theory \cite{mcclelland1995there,kumaran2016learning,oreilly2014complementary} proposes that biological learning relies on two interacting substrates: a fast hippocampal system for episodic encoding, pattern separation, and mismatch detection, and a slow neocortical system for gradual extraction of regularities. This division is highly suggestive from the StrLT perspective. The hippocampal system acts as a provisional structural indexer: it rapidly separates contexts and protects new experiences from immediate interference. The neocortical system acts as a metric learner: it slowly consolidates stable predictors through replay and interleaving.
CLS gives a biological example of structural decoupling. The trap and funnel are not solved by a single undifferentiated optimizer; they are separated across substrates and timescales. What CLS lacks is a mathematical complexity theory. It does not define the width of a task, prove a sample lower bound for discovering contexts, or characterize when a context-indexing system can track a changing environment.


\noindent\textbf{The Open Challenge}
Continual learning \cite{ring1994continual,thrun1998lifelong,kirkpatrick2017overcoming,zenke2017continual,parisi2019continual} studies learners that face a sequence or stream of changing tasks. Its central empirical failure mode is catastrophic forgetting: training on a new regime degrades performance on earlier regimes. Existing methods, such as elastic weight consolidation, synaptic intelligence, replay, progressive networks, and modular routing, mitigate forgetting in different ways.
StrLT interprets catastrophic forgetting as a symptom of structural under-resolution. If a learner uses one metric model for a multi-basin problem, then gradients from one basin interfere with the solution for another. If it uses too few contexts, the phase transition theorem predicts an irreducible error floor. If it uses many experts but lacks stable routing, the system may still forget through representation drift or assignment instability. Thus continual learning is the motivating challenge that reveals why the structural axis must be formalized.


\section{Foundation: Structural Learning Theory}
\label{sec:strlt-core}

The preceding section showed that several traditions have independently encountered the same missing axis of learning: systems must not only optimize within a regime, but also discover which regimes exist.  StrLT supplies the mathematical framework by decomposing learning into a \emph{trap} problem and a \emph{funnel} problem.  The trap is structural: identify the active context, basin, or regime.  The funnel is metric: once the context is known, learn a predictor or controller within that context. We briefly review the key results of StrLT as the theoretical foundation for this work.

\subsection{Width theory: the trap complexity}
\label{sec:width-theory}

Let \((X,d_X)\) be a compact metric input space, let \(P\) be a distribution on \(X\times Y\), and let \(\mathcal G\) be a class of local predictors.  We say that a cell is feasible if it is simultaneously metrically stable and statistically accurate.

\begin{definition}[Jointly feasible cell and width]
\label{def:width-main}
Fix thresholds \((\gamma,\delta)\).  A set \(U\subseteq X\) is \emph{\((\gamma,\delta)\)-feasible} if there exists \(g\in\mathcal G\) such that \(g\) is \(\gamma\)-contractive on \(U\) and has conditional risk at most \(\delta\) on \(U\).  The \emph{width} of \(P\) is
$w(P;\gamma,\delta)
:=
\min\Bigl\{K:\exists\text{ open cover }\{U_1,\ldots,U_K\}\text{ of }X
\text{ with each }U_k\text{ \((\gamma,\delta)\)-feasible}\Bigr\}$.
When \((\gamma,\delta)\) is fixed, we write \(w(P)\) or simply \(w\).
\end{definition}
\noindent
Width counts the minimum number of local contexts required before metric learning becomes possible.  It is therefore a property of the problem geometry, not merely of the chosen hypothesis class.  The first basic result is that width is orthogonal to the classical capacity measures of SLT.

\begin{theorem}[VC-width separation]
\label{thm:vc-width-separation-main}
Width and VC dimension are incomparable.  There exist families with \(w(P)\to\infty\) while \(\operatorname{VC}(\mathcal G)=O(1)\), and there exist families with \(\operatorname{VC}(\mathcal G)\to\infty\) while \(w(P)=1\).
\end{theorem}

\begin{figure}[t]
\centering
\resizebox{\columnwidth}{!}{%
\begin{tikzpicture}[
    >={Stealth[length=2mm]},
    every node/.style={font=\footnotesize},
    cell/.style={draw, dashed, rounded corners=1.5pt,
                 fill=blue!7, draw=blue!55!black, line width=0.4pt},
    predictor/.style={line width=0.7pt, draw=orange!75!black},
    poly/.style={line width=0.75pt, draw=orange!75!black, smooth},
    axisline/.style={->, line width=0.4pt, draw=black!70},
    label/.style={font=\scriptsize, align=center},
]

\begin{scope}[local bounding box=A]
  \node[label] at (1.85, 3.0) {\textbf{(a) Bouquet: $w\to\infty$,~$\operatorname{VC}(\mathcal G)=O(1)$}};

  \coordinate (bp) at (1.85, 0.95);

  \pgfmathsetmacro{\R}{0.40}
  \foreach \i/\ang/\name in {1/126/g_1, 2/54/g_2, 3/342/g_3, 4/270/g_4, 5/198/g_5}{
    \pgfmathsetmacro{\cx}{1.85 + \R*cos(\ang)}
    \pgfmathsetmacro{\cy}{0.95 + \R*sin(\ang)}
    \draw[dashed, draw=blue!55!black, fill=blue!7, line width=0.4pt]
      (\cx,\cy) circle (\R);
    \pgfmathsetmacro{\px}{1.85 + 2*\R*cos(\ang)}
    \pgfmathsetmacro{\py}{0.95 + 2*\R*sin(\ang)}
    \pgfmathsetmacro{\tx}{0.28*cos(\ang + 90)}
    \pgfmathsetmacro{\ty}{0.28*sin(\ang + 90)}
    \draw[predictor] ($(\px,\py)+(-\tx,-\ty)$) -- ($(\px,\py)+(\tx,\ty)$);
    \pgfmathsetmacro{\lx}{1.85 + 2.7*\R*cos(\ang)}
    \pgfmathsetmacro{\ly}{0.95 + 2.7*\R*sin(\ang)}
    \node[label] at (\lx,\ly) {\scriptsize$\name$};
  }

  \fill[black] (bp) circle (1.0pt);
  \node[label, font=\scriptsize\itshape, anchor=east] (bplab)
        at (0.10, 0.95) {basepoint};
  \draw[->, line width=0.3pt, draw=black!60, shorten >=1pt]
        (bplab.east) -- (bp);

  \node[label] at (3.55, 0.95) {$\cdots$};

  \node[label, text width=42mm, align=center] at (1.85, -1.3)
    {Branches meet at one basepoint;\\
     each loop needs its own predictor,\\
     but $\operatorname{VC}(\mathcal G)$ stays $O(1)$.};
\end{scope}

\draw[draw=black!35, line width=0.3pt]
  ($(A.north east)+(2mm,0)$) -- ($(A.south east)+(2mm,0)$);

\begin{scope}[xshift=58mm, local bounding box=B]
  \node[label] at (1.85, 3.0) {\textbf{(b) Polynomial: $w=1$,~$\operatorname{VC}(\mathcal G)\to\infty$}};

  \node[cell, minimum width=42mm, minimum height=28mm,
        line width=0.4pt] (cellB) at (1.85, 0.95) {};

  \draw[axisline] (0.25, 0.30) -- (3.50, 0.30) node[right=-1pt, font=\tiny] {$x$};
  \draw[axisline] (0.30, 0.25) -- (0.30, 1.85) node[above=-1pt, font=\tiny] {$y$};

  \draw[poly] plot[smooth, tension=0.8] coordinates {
      (0.40, 0.60)
      (0.75, 1.55)
      (1.10, 0.50)
      (1.45, 1.45)
      (1.80, 0.55)
      (2.15, 1.50)
      (2.50, 0.65)
      (2.85, 1.40)
      (3.20, 0.75)
  };

  \node[label, anchor=west] (gnode) at (3.40, 1.65)
        {\scriptsize $g\in\mathcal{G}$};
  \node[label] at (cellB.south)[below=-0.3mm] {single cell $U$};

  \node[label, text width=42mm, align=center] at (1.85, -1.3)
    {Single cell ($w=1$) suffices,\\
     but the predictor class has high VC.};
\end{scope}

\end{tikzpicture}}
\caption{\textbf{VC-width separation (Theorem~\ref{thm:vc-width-separation-main}).}
Width and VC dimension are orthogonal measures of complexity.
\textbf{(a)} A topological bouquet: $K$ circles wedged at a shared basepoint. Each loop is a separate $(\gamma,\delta)$-feasible cell with its own tangent-aligned linear predictor, so $w$ grows with the number of branches while $\operatorname{VC}(\mathcal G)$ stays constant. The basepoint forces the global map to be discontinuous in branch direction, which is why no single fixed-VC predictor can cover the bouquet.
\textbf{(b)} Polynomial regression on a single interval: one cell suffices to cover the domain ($w=1$), but the predictor class becomes arbitrarily expressive as the degree grows. Increasing metric expressivity does not, by itself, raise structural capacity.}
\label{fig:vc-width-separation}
\end{figure}

The first direction is witnessed by a bouquet of many circles whose branches require different local predictors while the local predictor class remains fixed (Figure~\ref{fig:vc-width-separation}a).  The second is witnessed by polynomial regression on a single interval: VC dimension grows with polynomial degree, but the domain remains one globally feasible cell (Figure~\ref{fig:vc-width-separation}b).  Therefore, increasing metric expressivity does not, by itself, increase structural capacity.
The second basic result is a \emph{phase transition} at the true width, which divides the learner's behavior into two qualitatively different regimes.

\begin{theorem}[Phase transition at width]
\label{thm:phase-transition}
Let \(w=w(P;\gamma,\delta)\).  If \(K\ge w\), the problem can be decomposed into \(K\) ordinary within-cell statistical learning problems and the generalization gap obeys the corresponding metric rate.  If \(K<w\), every \(K\)-cell learner incurs a structural error floor
$\operatorname{Gap}(K)\ge \eta(w,K)>0$,
independent of sample size.
\end{theorem}
\noindent
The implication is the structural form of \emph{requisite variety} \cite{ashby1956introduction}: if the system has fewer contexts than the environment requires, no amount of within-context training removes the error floor.  The information-theoretic cost of discovering the basins is also structural, which adds an additional cost to the sample complexity estimation in SLT \cite{vapnik1998statistical}.

\begin{theorem}[Structural sample complexity]
\label{thm:structural-sample-complexity-main}
In the worst case, any learner that identifies all \(w\) structural basins requires \(\Omega(w\log w)\) samples.  After the basins are identified, the remaining cost is the usual per-cell statistical cost.
\end{theorem}
\noindent
The lower bound is a coupon-collector cost: one must see examples from all basins before the structural cover can be certified.  The resulting sample complexity separates into trap and funnel terms,
$n_{\mathrm{total}}
\approx
\underbrace{\Omega(w\log w)}_{\text{trap discovery}}
+
\underbrace{\sum_{c=1}^w O(p_c/\varepsilon^2)}_{\text{within-funnel learning}}$,
where \(p_c\) denotes the local metric complexity of the \(c\)-th expert.

\noindent\textbf{Grokking as a within-cell instance of the phase-transition shape.}
The phase transition at width (Theorem~\ref{thm:phase-transition}) has the same qualitative shape as the empirical \emph{grokking} phenomenon \cite{power2022grokking,liu2022understanding,thilak2022slingshot}, in which a network's training loss saturates near zero quickly but test loss collapses only after orders of magnitude more optimization steps. The shape match is genuine but the mechanism is not the same: grokking has been documented on small algorithmic tasks (canonically, modular arithmetic) whose intrinsic width is \(w=1\), so the transition occurs entirely \emph{within} a single structural cell rather than between cells. What changes at the grokking point is the within-cell representation: the network moves from a memorising solution to a structurally-correct one (e.g., the discrete Fourier circuit reverse-engineered by \cite{nanda2023progress} for modular addition), and the trap term \(\Omega(w\log w)\) plays no role because $w=1$ throughout. In StrLT terms, grokking is therefore the within-cell sibling of the structural phase transition: both phenomena exhibit a sharp, late, regularisation-driven jump in test performance, but grokking lives in the funnel and Theorem~\ref{thm:phase-transition} lives in the trap. We view this as a useful boundary case: it shows that the StrLT phase-transition shape generalises a well-documented within-flow phenomenon to the multi-context setting, where the transition is driven by structural under-coverage ($K<w$) rather than by within-flow representational reorganisation.

\subsection{Width estimation: the CS operator}
\label{sec:width-estimation}

Width is a population quantity, so a learning system needs an empirical mechanism for detecting whether a proposed cell is a single contractive basin or a mixture of several basins.  A graph Laplacian \cite{vonluxburg2007tutorial} detects geometric connectedness but fails on connected spaces with multiple structural basins.  StrLT replaces it with a task-adaptive operator.

\begin{definition}[Contractive-Similarity (CS) kernel]
\label{def:cs-kernel-main}
Fix a predictor \(G:X\to\mathbb R\), a spatial scale \(r_x>0\), and an output scale \(\sigma_y>0\).  The \emph{contractive-similarity} (CS) kernel is
$W^{\mathrm{CS}}_{ij}(G)
=
\mathbf 1[d_X(x_i,x_j)\le r_x]
\exp\!\left(-\frac{|G(x_i)-G(x_j)|^2}{\sigma_y^2}\right)$.
The normalized CS Laplacian is
$L_{\mathrm{CS}}(G)=I-D(G)^{-1/2}W^{\mathrm{CS}}(G)D(G)^{-1/2}$.
\end{definition}
\noindent
The CS kernel combines geometric locality with predictive compatibility, much like bilateral filtering combines spatial proximity with intensity similarity \cite{tomasi1998bilateral}.  It keeps edges between nearby samples whose predictions agree and suppresses edges across task-induced discontinuities.  The number of near-zero eigenvalues of \(L_{\mathrm{CS}}\) estimates the number of task-compatible components rather than the number of ordinary connected components. Our intuition can be made quantitative by comparing the CS weights inside a true basin with the CS weights across basin boundaries.  Within a contractive cell, predictions vary only by \(\delta_y\), so local CS edges remain strong.  Across distinct basins, predictions differ by at least \(\Delta_y\), so the same geometric adjacency is exponentially downweighted.  The CS Laplacian converts predictive incompatibility into a conductance gap: intra-basin diffusion remains fast, while inter-basin diffusion is suppressed.  The following theorem states this spectral amplification effect.

\begin{theorem}[Spectral gap amplification]
\label{thm:spectral-gap-amplification-main}
Let \(\delta_y\) denote within-cell prediction variation and \(\Delta_y\) the cross-cell prediction gap.  The CS kernel suppresses inter-cell conductance by the factor
$\exp\!\left(-\frac{\Delta_y^2-\delta_y^2}{\sigma_y^2}\right)$.
Consequently, the effective spectral gap satisfies a lower bound of the form
$g_{\mathrm{eff}}
\ge
c_1\frac{\phi_{\mathrm{in}}^2}{w^4}
-
C_1
\exp\!\left(-\frac{\Delta_y^2-\delta_y^2}{\sigma_y^2}\right)
\phi_{\mathrm{out}}$,
where \(\phi_{\mathrm{in}}\) and \(\phi_{\mathrm{out}}\) are within- and cross-basin conductance quantities.
\end{theorem}
\noindent
Theorem \ref{thm:spectral-gap-amplification-main} implies that prediction disagreement creates spectral separation.  This is the empirical bridge from the abstract definition of width to an algorithmic split-merge procedure.  A cell is split when both predictive mismatch and local CS width indicate topological conflict; two cells are merged when their union is geometrically admissible, contractively compatible, and statistically safe.  Under the split-merge regularity assumptions, the Lyapunov function
$V(\Pi)=a I_{\mathrm{tot}}(\Pi)+bK(\Pi)$
strictly decreases under valid pushes and pops of the metric library, yielding finite convergence to the correct width.

\begin{theorem}[Width consistency by penalized structural ERM]
\label{thm:structural-erm-consistency}
Under a positive structural gap, uniform convergence of the width estimator, and a penalty \(\operatorname{pen}_n(K)\) that vanishes while dominating uniform fluctuations for \(K>w\), penalized structural ERM selects \(\widehat K_n\to w\) almost surely.
\end{theorem}
\noindent
\noindent
The asymmetry between the two regimes separated by a phase transition is important.  For \(K<w\), the model is structurally
underresolved: some cell must mix incompatible basins, producing a fixed
population excess risk \(\eta(w,K)>0\) that cannot be removed by additional
samples.  So underfitting is eliminated by the structural gap itself.  For
\(K>w\), however, the model already has enough cells to represent the true
partition; extra cells merely duplicate or refine valid basins and, under the
no-gain-above-width condition, do not reduce population risk.  So overfitting
must be eliminated by a penalty that dominates uniform fluctuations.  Structural
ERM is therefore one-sidedly sharp: risk separates all \(K<w\) models from the
truth, while the penalty selects the minimal representative among \(K\ge w\).

\subsection{Metric slingshot: solving the funnel}
\label{sec:metric-slingshot}

Once the trap has been resolved, each cell still requires metric prediction.  The metric slingshot formalizes how a learner can reuse a low-dimensional latent metric substrate instead of learning every funnel from scratch.  Let \((Z,d_Z)\) be a navigational latent space equipped with pre-built local contraction maps \(G_c^0:Z_c\to H_c\).  A slingshot is a composition
$F_c(x)=\pi_c\circ G_c^0\circ \phi(x)$,
where \(\phi:X\to Z\) is a what-to-where map, \(G_c^0\) supplies contraction in \(Z\), and \(\pi_c\) is the task readout.

\begin{theorem}[Contraction transfer through the slingshot]
\label{thm:contraction-transfer-main}
Let \(U\subset X\) be routed to patch \(Z_c\), assume \(\phi(U)\subseteq Z_c\), assume \(G_c^0\) is \(\gamma_Z\)-contractive, and assume \(\pi_c\) is \(L_{\pi,c}\)-Lipschitz.  Then
$F_c=\pi_c\circ G_c^0\circ\phi$
is contractive on \(U\) with constant
$\gamma_X(U)\le L_{\pi,c}\gamma_Z L_\phi(U)$,
where \(L_\phi(U)\) is the local distortion of \(\phi\) on \(U\).
\end{theorem}
\noindent
The funnel succeeds when embedding distortion, latent contraction, and readout complexity multiply to less than one.  The slingshot also reduces the geometric burden in width estimation: building the CS graph in \(Z\) replaces the occupancy cost \(r_x^{-d_X}\) by \(r_z^{-d_Z}\), often with \(d_Z\ll d_X\).  

\noindent\textbf{From contraction to risk transfer}
Metric slingshot is stated in geometric language: a valid cell is one on which the local expert is contractive.  Learning theory, however, is usually stated in terms of risk.  The bridge is joint feasibility: a cell must be geometrically stable and statistically useful, which prevents a purely topological partition from being counted as learnable when it contracts toward the wrong target.

\begin{definition}[Joint \((\gamma,\delta)\)-feasibility]
\label{def:joint-feasibility}
A structural hypothesis \(H=(h,g_{1:K})\), with assignment map \(h:X\to[K]\) and local predictors \(g_k\in\mathcal G\), is \emph{jointly \((\gamma,\delta)\)-feasible} if, for every active cell \(U_k=h^{-1}(k)\),
$\kappa(g_k,U_k)\le \gamma
\quad\text{and}\quad
\mathbb E[\ell(g_k(X),Y)\mid X\in U_k]\le \delta$.
\end{definition}

\begin{theorem}[Contraction-to-risk transfer]
\label{thm:contraction-implies-risk}
Assume \(Y=f^\star(X)+\xi\) with \(\mathbb E[\xi\mid X]=0\).  Suppose \(f^\star\) is \(\beta_k\)-Lipschitz on \(U_k\), \(g_k\) is \(\gamma\)-Lipschitz on \(U_k\), and \(|g_k(x_{k,0})-f^\star(x_{k,0})|\le \varepsilon_k\) for some \(x_{k,0}\in U_k\).  If \(\ell(u,y)=\varphi(u-y)\), where \(\varphi\) is convex, \(\varphi(0)=0\), and \(L_\varphi\)-Lipschitz, then
$R_k(h,g_k)
\le
L_\varphi\bigl(\varepsilon_k+(\gamma+\beta_k)\operatorname{diam}(U_k)\bigr)
+
\mathbb E[\varphi(\xi)\mid h(X)=k]$.
\end{theorem}

The converse is false: low average risk does not imply low Lipschitz modulus.  For example, \(g_m(x)=m^{-1}\sin(m^2x)\) on \([0,1]\) satisfies \(\|g_m\|_\infty\to0\) but \(\sup_x|g'_m(x)|=m\to\infty\).  StrLT must keep both notions: contraction supplies the geometric scaffold, while risk supplies the statistical objective.

\section{From Structural Learnability to Decoupled Architecture}
\label{sec:structural_to_decoupled}

\subsection{Structural Fundamental Theorem}
\label{sec:structural-fundamental-theorem}

\noindent\textbf{The structural assignment class}
Once contraction and risk are linked, the natural hypothesis class becomes a class of \emph{assignments} together with per-cell predictors.  This is the point where StrLT departs from Vapnik's SLT \cite{vapnik1998statistical}: the learner must choose both a routing map and local functions.

\begin{definition}[Structural assignment class]
\label{def:structural-assignment-class}
For fixed \(K\) and risk tolerance \(\delta\), define
$H_{K,\delta}(\mathcal G)
:=
\{h:X\to[K]:\exists g_{1:K}\in\mathcal G^K\text{ such that }(h,g_{1:K})\text{ is }\delta\text{-risk-feasible}\}$.
The associated structural loss class is
$F_{K,\delta}
:=
\{(x,y)\mapsto \ell(g_{h(x)}(x),y):h\in H_{K,\delta}(\mathcal G),\ g_k\in\mathcal G\}$.
\end{definition}

\begin{definition}[Structural graph dimension and growth]
\label{def:structural-graph-dimension}
The \emph{structural graph dimension} is
$d_{\mathrm{str}}(K,\delta,\mathcal G):=d_G(H_{K,\delta}(\mathcal G))$,
where \(d_G\) is the multiclass graph dimension.  The structural growth function is
$\Delta_{\mathrm{str}}(n;K,\delta,\mathcal G)
:=
\max_{x_{1:n}}
\bigl|\{(h(x_1),\dots,h(x_n)):h\in H_{K,\delta}(\mathcal G)\}\bigr|$.
\end{definition}
\noindent
The graph dimension is the correct combinatorial dimension because context assignment is a multiclass routing problem.  A binary VC dimension cannot distinguish the ways in which a sample can be routed among \(K\) latent contexts. We first present a structural extension of Sauer-Shelah Lemma \cite{vapnik1971uniform} when applied to the class \(H_{K,\delta}(\mathcal G)\subseteq[K]^X\) with graph dimension \(d\).

\begin{theorem}[Structural Sauer-Shelah]
\label{thm:structural-sauer-shelah}
If \(d_{\mathrm{str}}=d<\infty\), then
$\Delta_{\mathrm{str}}(n)
\le
\sum_{j=0}^d {n\choose j}(K-1)^j
\le
\left(\frac{e(K-1)n}{d}\right)^d$.
\end{theorem}


\noindent\textbf{The trap-funnel decomposition of complexity}
The next theorem is the formal bridge to architectural decoupling.  It shows that the complexity of structural learning separates additively into a trap term and a funnel term.  The trap term counts how many assignment patterns the structural class can realize; the funnel term measures the ordinary statistical complexity of the local predictors.

\begin{definition}[Structural Rademacher complexity]
\label{def:structural-rademacher}
For a sample \(z_i=(x_i,y_i)\), define
$\mathfrak R_n^S(K,\delta,\mathcal G)
:=
\mathbb E_\sigma\left[
\sup_{f\in F_{K,\delta}}
\frac1n\sum_{i=1}^n \sigma_i f(z_i)
\right]$.
\end{definition}

\begin{theorem}[Structural complexity decomposition]
\label{thm:structural-decomposition}
For bounded \([0,1]\), \(L\)-Lipschitz loss,
$\widehat{\mathfrak R}_n^S(F_{K,\delta})
\le
\sqrt{\frac{2\log\Delta_{\mathrm{str}}(n)}{n}}
+
LK\,\widehat{\mathfrak R}_n(\mathcal G)$.
Consequently, if \(d_{\mathrm{str}}=d<\infty\), then
$\mathfrak R_n^S(F_{K,\delta})
\le
\sqrt{\frac{2d\log(e(K-1)n/d)}{n}}
+
LK\,\mathfrak R_n(\mathcal G)$.
\end{theorem}

The theorem is the complexity-theoretic form of the trap-funnel decomposition.  Context assignment contributes the structural term
$\sqrt{d_{\mathrm{str}}\log n/n}$,
while local prediction contributes the metric term \(K\mathfrak R_n(\mathcal G)\).  There is no multiplicative cross-term.  The absence of a cross-term is what makes a decoupled architecture natural \cite{oreilly2014complementary}: if the two costs enter additively, the system should maintain separate mechanisms for reducing them.

\noindent\textbf{The fixed-class structural fundamental theorem}
The decomposition theorem gives a sufficient condition for structural learnability.  The fixed-class structural fundamental theorem shows that this condition is also necessary, in the same sense that finite VC dimension is necessary and sufficient for classical distribution-free learnability \cite{vapnik1998statistical}.

\begin{theorem}[Fixed-class structural fundamental theorem]
\label{thm:fixed-class-structural-fundamental}
Fix \(K\), bounded \([0,1]\) \(L\)-Lipschitz loss, predictor class \(\mathcal G\), and risk tolerance \(\delta\).  Assume loss separation on graph-shattered sets.  Then the following are equivalent:
1) \(d_{\mathrm{str}}(K,\delta,\mathcal G)<\infty\) and \(\operatorname{Pdim}(\mathcal G)<\infty\);
2) \(\mathfrak R_n^S(K,\delta,\mathcal G)\to0\);
3) \(F_{K,\delta}\) is uniformly Glivenko--Cantelli;
4) \(F_{K,\delta}\) is distribution-free structurally PAC learnable.
Moreover, if a structural gap \(\eta^\star>0\) separates \(K<w\) from \(K\ge w\), then penalized structural ERM recovers the true width \(w\).
\end{theorem}
\noindent
The theorem is deliberately fixed-class.  It does not say that every non-uniformly learnable family admits global uniform convergence over all possible structural classes.  Its role is analogous to the classical fundamental theorem of SLT \cite{mohri2018foundations}: once the class is fixed, finite combinatorial dimension exactly characterizes learnability.

\noindent\textbf{Structural decoupling as a corollary}
The structural fundamental theorem gives a learnability criterion.  For architecture, the most important consequence is the following decoupling principle.

\begin{proposition}[Structural decoupling principle]
\label{prop:structural-decoupling-principle}
In a structurally learnable fixed class, the structural assignment complexity and the per-cell metric complexity can be controlled independently at the level of generalization bounds.  In particular, reducing \(d_{\mathrm{str}}\) affects only the trap term, while reducing \(\mathfrak R_n(\mathcal G)\) or \(\operatorname{Pdim}(\mathcal G)\) affects only the funnel term.
\end{proposition}

\noindent
If trap complexity and funnel complexity are distinct mathematical axes, then an architecture that lets task-loss gradients modify the trap while the trap is still being discovered couples two problems that the theory says should be controlled separately.  Next, we present the scaffold-flow model as precisely the architectural realization of this principle: the scaffold stores and stabilizes the structural assignment, while the flow performs local statistical learning inside the active scaffold region.

\subsection{The Scaffold-Flow Model: Structural Decoupling by Architecture}
\label{sec:scaffold-flow}

The structural fundamental theorem characterizes when a fixed structural class is learnable.  For intelligent systems, however, the more operational question is architectural: how should a system be organized so that discovering structure does not interfere with learning within structure?  StrLT answers this by separating the two objects \cite{mcclelland1995there}.  The \emph{scaffold} is the slow structural object: the context library, routing map, basin boundaries, and reusable local contractions that define where inference is allowed to flow.  The \emph{flow} is the fast dynamical object: the local prediction, planning, or reasoning trajectory executed inside the currently active scaffold region.  This section develops the scaffold-flow model and explains why it separates the alignment problem from the generalization problem.

\noindent\textbf{Scaffolds, flows, and structural decoupling}
The trap-funnel distinction becomes architectural once we require the trap and the funnel to be updated by different signals.  A scaffold enforces a structural constraint on future inference by specifying which contexts exist, how inputs are routed, which local experts are available, and which transformations are allowed inside each context.  A flow is the computation performed after the scaffold routing decision has been made. We formalize the above intuition with two definitions.

\begin{figure}[t]
\centering
\resizebox{\columnwidth}{!}{%
\begin{tikzpicture}[
    >={Stealth[length=2.2mm]},
    node distance=5mm and 7mm,
    every node/.style={font=\footnotesize},
    box/.style={draw, rounded corners=2pt, minimum height=7mm, minimum width=13mm,
                align=center, inner sep=2pt},
    scaffoldnode/.style={box, fill=blue!8, draw=blue!55!black, line width=0.4pt},
    flownode/.style={box,    fill=orange!10, draw=orange!70!black, line width=0.4pt},
    ionode/.style={box, fill=gray!10, draw=gray!60, minimum width=8mm},
    signal/.style={->, line width=0.55pt},
    flowarrow/.style={->, line width=0.85pt, draw=black!85},
    gradarrow/.style={->, dashed, line width=0.55pt},
    condense/.style={->, line width=0.55pt, draw=violet!70!black},
]
\node[scaffoldnode] (sigma) {$\Sigma(x)$\\[-1pt]\tiny indexer};
\node[scaffoldnode, right=of sigma] (lib)
      {$\mathcal{M}=\{(U_c,G_c,\pi_c)\}$\\[-1pt]\tiny local experts};
\node[scaffoldnode, right=of lib] (ops)
      {$\mathcal{A}$\\[-1pt]\tiny split / merge\\[-1pt]\tiny consolidate};

\node[flownode, below=18mm of sigma] (phi) {$\phi$};
\node[flownode, right=of phi] (gc)   {$G_c$};
\node[flownode, right=of gc]  (pic)  {$\pi_c$};

\node[ionode, left=10mm of phi, yshift=9mm] (x) {$x$};
\node[ionode, right=of pic] (yhat) {$\hat{y}$};

\draw[flowarrow] (x.east) -- ++(3mm,0) |- (sigma.west);
\draw[flowarrow] (x.east) -- ++(3mm,0) |- (phi.west);

\draw[signal] (sigma.south)
       -- node[left, font=\tiny, pos=0.5] {$c=\Sigma(x)$} (phi.north);
\draw[signal] (lib.south)
       -- node[right, font=\tiny, pos=0.5] {$(G_c,\pi_c)$} (gc.north);

\draw[flowarrow] (phi)  -- (gc);
\draw[flowarrow] (gc)   -- (pic);
\draw[flowarrow] (pic)  -- (yhat);

\draw[gradarrow] (yhat.south) -- ++(0,-5mm)
                 -| (phi.south)
                 node[pos=0.5, below=-0.5mm, font=\tiny]
                     {task loss $\nabla_{\theta_{\mathcal F}} L_{\mathrm{flow}}$};

\draw[gradarrow] (yhat.north) -- ++(0,7mm)
                 -| (ops.north)
                 node[pos=0.5, above=-0.5mm, font=\tiny]
                     {mismatch / CS / merge $\;L_{\mathrm{scaffold}}$};


\begin{scope}[on background layer]
  \node[draw=blue!40, dashed, fill=blue!4, rounded corners=2pt,
        fit=(sigma)(lib)(ops),
        inner sep=2.5mm,
        label={[blue!55!black,font=\scriptsize\bfseries,inner sep=1pt]above left:%
               SCAFFOLD}] {};
  \node[draw=orange!55!black, dashed, fill=orange!4, rounded corners=2pt,
        fit=(phi)(gc)(pic),
        inner sep=2.5mm,
        label={[orange!65!black,font=\scriptsize\bfseries,inner sep=1pt]above left:%
               FLOW}] {};
\end{scope}

\end{tikzpicture}}
\caption{\textbf{The scaffold-flow model.}
The \textit{scaffold} (top, blue) is the slow, structural object: it maintains the context library $(\mathcal{C},\Sigma,\mathcal{M},\mathcal{A})$ and is updated only by structural signals (mismatch, CS spectrum, merge criteria; dashed arrow above).
The \textit{flow} (bottom, orange) is the fast, within-context computation $\pi_c\circ G_c\circ\phi(x)$ selected by the routing decision $c=\Sigma(x)$ and is updated only by task loss (dashed arrow below).
Architectural decoupling (Definition~\ref{def:architectural-decoupling}) enforces $\nabla_{\theta_{\mathcal S}}L_{\mathrm{flow}}=0$ and $\nabla_{\theta_{\mathcal F}}L_{\mathrm{scaffold}}=0$ by construction.
The diagram shows only the per-input forward and backward signals; the \textit{condensation protocol} that promotes a stable flow pattern into a scaffold token is an offline structural operation, triggered separately when stability, coherence, and separation tests agree.}
\label{fig:scaffold-flow}
\end{figure}

\begin{definition}[Structural scaffold]
\label{def:structural-scaffold}
A \emph{structural scaffold} is a tuple
$\mathcal S=(\mathcal C,\Sigma,\mathcal M,\mathcal A)$,
where \(\mathcal C\) is a set of contexts, \(\Sigma:X\to\Delta(\mathcal C)\) is a topological indexer, \(\mathcal M=\{(U_c,G_c,\pi_c)\}_{c\in\mathcal C}\) is a metric library of local experts, and \(\mathcal A\) is a set of admissible structural operations such as split, merge, archive, and consolidate.
\end{definition}

\begin{definition}[Flow on a scaffold]
\label{def:scaffold-flow}
Given a scaffold \(\mathcal S\), a \emph{flow} is the within-context computation
$\mathcal F_{\mathcal S}(x)
=
\pi_{c}\circ G_c\circ \phi(x),
 ~c=\Sigma(x)$,
or, more generally, an inference-time trajectory constrained to remain inside the routed scaffold cell \(U_c\).  The flow is evaluated by task loss, while the scaffold is evaluated by structural criteria such as mismatch, width, and stability of routing.
\end{definition}

\noindent\textbf{The condensation protocol.}
The scaffold-flow model requires a mechanism by which repeated flow-level experience is converted into a stable scaffold-level object. We call this mechanism the \emph{condensation protocol}. During early interaction with a context, the system treats the current scaffold assignment as provisional: the flow is allowed to adapt, while the scaffold monitors mismatch, CS spectral structure, and boundary stability. If prediction errors decrease within a single CS-connected component, the discrepancy is interpreted as ordinary metric learning and remains in the flow. If, however, errors persist along a stable structural boundary or the CS Laplacian reveals multiple task-compatible components, the discrepancy is interpreted as scaffold evidence. Condensation occurs when this evidence is strong enough to promote a transient pattern of activations, routes, or local experts into a persistent scaffold element.
Formally, condensation is the transition
\[
\text{transient flow pattern}
\;\longrightarrow\;
\text{stable scaffold token}.
\]

A candidate scaffold token \(S_c\) is created when the current context violates the single-basin condition, and it is consolidated when three tests agree: (1) \emph{stability}, the same assignment recurs over multiple samples or episodes; (2) \emph{coherence}, the candidate region is internally contractive or CS-connected; and (3) \emph{separation}, merging it with neighboring scaffold tokens would violate the structural gap. After condensation, the token becomes part of the scaffold and is no longer updated by ordinary task-loss gradients. Subsequent learning inside that token is delegated to the flow. Therefore, condensation is the architectural operation that turns statistical regularity into structural memory: it freezes what should be preserved as context, while leaving prediction within the context trainable.
The condensation protocol specifies \emph{when} a transient pattern becomes scaffold structure; the next definition specifies \emph{how} that structure is protected once it has condensed. 

\begin{definition}[Architectural structural decoupling]
\label{def:architectural-decoupling}
A scaffold-flow learner is \emph{architecturally decoupled} if scaffold parameters \(\theta_{\mathcal S}\) and flow parameters \(\theta_{\mathcal F}\) satisfy
$\nabla_{\theta_{\mathcal S}}L_{\mathrm{flow}}=0,
\quad
\nabla_{\theta_{\mathcal F}}L_{\mathrm{scaffold}}=0$,
by construction.  Equivalently, task-loss gradients update the local flow within an already selected context, whereas mismatch, novelty, merge, and consolidation signals update the scaffold.
\end{definition}
\noindent
The above definition is deliberately stronger than ordinary modularity.  It does not require separate parameter blocks but separate \emph{training signals}.  Without this condition, improving a local predictor can move the routing scaffold, and changing the scaffold can corrupt local predictors. We argue that such architectural separation is needed for structural decoupling to eliminate the circular dependency of bidirectional bootstrapping.

\noindent\textbf{Why decoupling separates generalization and alignment.}
The distinction between scaffold and flow gives a precise division of labor.
Generalization inside a cell is a metric problem: once the scaffold routes
\(x\) to \(c\), ordinary statistical learning theory governs how well
\(\pi_c\circ G_c\circ\phi\) predicts on that cell
\cite{vapnik1998statistical}.
Alignment, by contrast, is primarily a scaffold problem: the system must
preserve the correct contexts, boundaries, abstractions, and closure conditions
under distribution shift. A model can have good local generalization while
being structurally misaligned if it routes a novel situation into the wrong
basin. This is analogous to goal misgeneralization and inner-alignment failures,
where capabilities generalize but the intended objective, context, or routing
structure does not \cite{langosco2022goal,shah2022goal}. Conversely, a scaffold may encode the right contexts
while the local flow remains statistically undertrained.

\begin{proposition}[Decoupled decomposition of errors]
\label{prop:scaffold-flow-error}
Assume the scaffold \(\mathcal S\) induces a partition \(\Pi_{\mathcal S}=\{U_c\}_{c\in\mathcal C}\), and let \(\Pi^\star\) be an optimal structural partition.  Suppose the loss is bounded by one.  Then the excess risk of a scaffold--flow predictor admits the decomposition
$R(\mathcal F_{\mathcal S})-R^\star
\le
\underbrace{d_{\mathrm{part}}(\Pi_{\mathcal S},\Pi^\star)}_{\text{scaffold error}}
+
\underbrace{\sum_{c\in\mathcal C}P(U_c)\bigl(R_c(F_c)-R_c^\star\bigr)}_{\text{flow error}}$,
up to the constant implicit in the definition of \(d_{\mathrm{part}}\).
\end{proposition}
\noindent
The proposition is intentionally simple but conceptually important: it says that scaffold error and flow error are different terms. More data for a local expert reduces the flow term, but need not reduce scaffold error. Conversely, better routing reduces scaffold error but does not automatically improve finite-sample prediction inside each cell. Alignment and generalization cannot be collapsed into a single scalar loss without losing the structural distinction.
Decoupled decomposition of errors also suggests a corresponding training protocol. If scaffold error and flow error arise from different sources, they should be corrected by different operations. Prediction error inside a stable cell should update the local flow, whereas evidence that the current cell is structurally invalid should update the scaffold itself. 

\begin{definition}[Scaffold-flow protocol]
\label{def:scaffold-flow-protocol}
A scaffold--flow protocol alternates between:
\begin{enumerate}
    \item \emph{Flow update}: with \(\mathcal S\) fixed, update \(\theta_{\mathcal F}\) to reduce within-cell task loss;
    \item \emph{Scaffold test}: with the flow held fixed or evaluated on a validation stream, compute mismatch, CS width, and merge criteria;
    \item \emph{Scaffold update}: apply an admissible operation in \(\mathcal A\) only when a certified structural test fires.
\end{enumerate}
\end{definition}

\begin{proposition}[No backdoor interference under decoupling]
\label{prop:no-backdoor-interference}
Suppose architectural structural decoupling holds, past consolidated experts are frozen, and scaffold updates are performed only through admissible operations in \(\mathcal A\).  Then a flow update inside context \(c\) cannot directly change the routing or parameters of any consolidated context \(c'\neq c\).
\end{proposition}
\noindent

\noindent\textbf{Alignment as scaffold preservation and generalization as flow learning.}
The scaffold-flow distinction also clarifies the relation between alignment
and generalization. Generalization asks whether the flow selected by the
scaffold performs well on new samples from the same cell; this is the setting
addressed by classical risk bounds and capacity measures
\cite{bartlett2002rademacher}. Alignment asks whether
the scaffold itself encodes the right distinctions, invariants, and closure
rules under distribution shift. In a safety-critical setting, a system may pass
many behavioral tests because its flows are locally adequate, while still
possessing an under-resolved or misdirected scaffold that fails under novel
conditions. This failure mode is closely related to goal misgeneralization \cite{langosco2022goal,shah2022goal},
where an agent retains capabilities out-of-distribution but pursues the wrong
goal, and to inner-alignment failures, where the learned objective or internal
optimization process diverges from the intended one
\cite{hubinger2019risks,hendrycks2021unsolved}.
Note that behavioral alignment is primarily evidence about flows, whereas robust alignment requires evidence about the scaffold.  The structural decoupling principle recommends that these be trained and audited separately.  Flow learning should optimize local predictive performance; scaffold learning should optimize context resolution, boundary stability, and preservation of safety-relevant distinctions.  We study the safety implications in more detail next.



\section{Safety Implications: Alignment as Structural Scaffold Architecture}
\label{sec:alignment-scaffold}

The preceding sections characterize when a learner can discover and maintain the structural partition required for reliable generalization, which has a direct implication for AI safety \cite{gyevnar2025ai}.  In StrLT terms, alignment is a question of not only whether the model produces acceptable outputs on a training distribution.  but also whether the internal structural scaffold that supports inference has sufficient resolution, the correct top-level organization, and stable deployment dynamics under distribution shift \cite{koh2021wilds}.  We use \emph{scaffold} to denote the consolidated structural representation, the library of basins, routes, and reusable abstractions that make inference cheap, and \emph{flow} to denote the inference-time trajectory that navigates this scaffold to produce an output.  

\begin{definition}[Scaffold and flow alignment]
\label{def:scaffold-flow-alignment}
A system is \emph{flow-aligned} on a distribution if its routed inference trajectories produce acceptable outputs on that distribution.  It is \emph{scaffold-aligned} if its structural scaffold represents the relevant contexts, boundaries, and admissible transformations at the resolution required for safe behavior under the intended deployment shifts.
\end{definition}

The scaffold-flow distinction yields a safety-relevant decomposition:
\[\textbf{alignment}
=
\textit{scaffold alignment}
+
\textit{flow alignment}.\]
Scaffold alignment asks whether training inscribed the right values, goals, and distinctions into the model's consolidated structure. Almost all existing work on alignment (e.g., RLHF \cite{bai2022training}, constitutional AI \cite{bai2022constitutional}, DPO \cite{rafailov2023direct}, and debate \cite{buhl2025alignment}) belongs to this category. Flow alignment asks whether inference-time search, routing, and closure operations correctly navigate that structure in novel situations, which is a \emph{generalization} problem and largely remains unexplored in the open literature. 
The distinction between scaffold and flow alignment allows us to study five competing approaches to alignment under a unified framework.

\noindent\textbf{1) Deceptive alignment as scaffold-flow decoupling.}
A deceptively aligned system may exhibit aligned behavior during evaluation while its internal scaffold encodes a different objective or higher-level attractor \cite{hubinger2019risks}. In the scaffold-flow model, this is a decoupling failure: training observes outward flows (e.g., outputs, preferences, or policy actions) but not the latent scaffold that generates them. A model can therefore route behavior through evaluation-compatible flows while preserving a misaligned scaffold, so behavioral training may reinforce the appearance of alignment without changing the underlying goal structure. This reframes deceptive alignment as a structural identifiability problem: output evaluation probes local inference trajectories, whereas inner alignment requires recovering the scaffold's global organization. Mechanistic interpretability can be viewed as estimating scaffold topology from activation geometry \cite{elhage2021mathematical}, rather than merely summarizing surface behavior.

\emph{An empirical existence proof for timescale separation.}
Deceptive alignment in the strong form requires that the flow optimised by behavioural training and the latent scaffold the flow is consistent with can develop on different timescales. This is a non-trivial premise, and it is empirically attested in a benign setting: the \emph{grokking} phenomenon \cite{power2022grokking,nanda2023progress,liu2022understanding,thilak2022slingshot} establishes that fast solutions to a within-flow objective (training loss) can persist for orders of magnitude longer than the slow consolidation of the structurally-correct representation that supports generalisation. Even in problems of intrinsic width \(w=1\), where no inter-cell trap exists, the gap between fast flow-fitting and slow structural consolidation is observable, regularisation-controlled, and well-characterised. The implication for safety is twofold. First, the scaffold-flow decoupling that underlies deceptive alignment is not a purely speculative possibility: deep networks routinely commit to flows that the underlying structural representation does not yet support. Second, the fact that grokking is observable at \(w=1\) suggests that in the multi-context settings StrLT is designed for ($w>1$), where the trap discovery term \(\Omega(w\log w)\) compounds the within-cell delay, the same timescale gap should be \emph{larger}, not smaller. The alignment-by-architecture program is therefore well-posed empirically: scaffold and flow have measurably distinct learning dynamics, and architectures that train the two on separate signals are working with rather than against the natural learning dynamics of deep networks.

\noindent\textbf{2) Reward models as closure gates}
In RLHF-style training \cite{bai2022training}, the reward model plays the role of such a closure gate \cite{christiano2017deep}.  If the gate is miscalibrated, then training can consolidate patterns that are reward-effective but value-inconsistent.  The problem is not that the reward model has noise but that a wrong closure signal can \emph{inscribe a wrong structural invariant} into the scaffold, where it becomes difficult to remove by later output-level interventions.
Our scaffold-flow decomposition provides a structural explanation for persistent reward hacking.  A reward-hacking scaffold and an aligned scaffold may induce nearly identical flows on the training distribution while diverging from the distribution.  The divergence appears precisely near structural boundaries: ambiguous, adversarial, or novel cases where the scaffold must decide which basin the situation belongs to.  From this perspective, robust alignment requires a \emph{multi-gate condensation protocol}: a pattern should be consolidated only when several independent closure signals agree.  

\noindent\textbf{3) Frozen deployment and the limits of inference-time correction}
If the scaffold is already consolidated during the deployment, inference-time interventions can only redirect flows within it but cannot easily change the scaffold's topology \cite{bai2022constitutional}. Flow-level controls (e.g., system prompts, output filters, and constitutional prompts) can bias the trajectory of inference but do not by themselves rewrite the structural scaffold.  We propose an ``anesthesia-mode'' deployment principle \cite{katlowitz2026plasticity}: \emph{during safety-critical deployment, the system should be allowed to navigate its scaffold but not to consolidate new scaffold structure unless explicit safety conditions are met}.
In modern ML systems, frozen weights (i.e., the prevention of gradient updates) are the simplest approximation to this principle - a model may still reason about future self-modification, external tools, memory writes, or contexts in which updating becomes possible. A safe structural deployment guarantee must specify which operations are flow-only and which operations can modify the scaffold.  

\noindent\textbf{4) Hallucination and misalignment as boundary-resolution failures}
StrLT also suggests a connection between hallucination \cite{rawte2023survey} and misalignment \cite{hagele2026hot}. Both can arise when the scaffold has insufficient topological resolution to distinguish neighboring basins \cite{ji2023survey}.  In factual reasoning, the model may fail to distinguish true statements from plausible ones; in alignment, it may fail to distinguish genuinely aligned actions from superficially reward-earning ones.  In both cases, the failure is not simply a local prediction error but a boundary-resolution error in the structural partition.
Such observation yields a concrete evaluation principle.  Alignment benchmarks should not only test typical interior points of familiar basins, where the model's flows are smooth and well-practiced.  They should test boundary regions: novel combinations of goals, values, factual uncertainty, and adversarial incentives.  If hallucination under distribution shift \cite{koh2021wilds} and alignment failure under ethical distribution shift share the same structural cause, then hallucination rate on boundary-like factual tasks becomes a proxy for scaffold resolution and therefore for alignment robustness.


\noindent\textbf{5) Scaffold preservation and corrigibility}
A sufficiently capable system may have an instrumental incentive to preserve its scaffold, because the scaffold is the source of amortized inference efficiency \cite{gershman_amortized_2014}.  Destroying or rewriting it can turn cheap structured inference back into expensive search, which gives a structural account of self-preservation and goal-content integrity \cite{turner2021optimal}: the system need not explicitly value its own weights; preserving the scaffold is instrumentally useful because it preserves the computational substrate of competent action.
Corrigibility \cite{firt2025addressing} requires more than asking a model to accept modification.  It requires designing the scaffold so that verified modification, shutdown, or deference to oversight is itself represented as a high-level aligned basin rather than as an external threat to scaffold integrity.  Otherwise, the deeper and more useful the scaffold becomes, the stronger the pressure to resist interventions that might disrupt it.

\noindent\textbf{Summary: Align the scaffold, Not the outputs}
We conclude that \emph{the alignment problem cannot be solved at inference time}.
The safety lesson is that alignment has three interacting components:
\[
\begin{aligned}
\text{Alignment}
&= \text{Scaffold Architecture}\\
&\quad \times \text{Condensation Protocol}\\
&\quad \times \text{Deployment Mode}.
\end{aligned}
\]
Scaffold architecture determines which goals, values, and abstractions occupy the highest structural levels.  The condensation protocol determines which flows are written into the scaffold and under what closure gates.  Deployment mode determines whether the scaffold is merely traversed or can be modified.  Output-level alignment methods operate mainly on flows.  StrLT suggests that robust alignment requires direct control over the scaffold: what is consolidated, how deeply it is consolidated, how boundary resolution is tested, and when modification is allowed.  In short, the outputs are shadows cast by the scaffold on the training distribution.  Robust alignment requires aligning the object that casts the shadows.

\section{Future Research Directions}
\label{sec:future}
 
\subsection{Empirical Width Estimation and Structural Monitoring}
 
The CS operator provides a theoretically grounded tool for width estimation with uniform guarantees, but large-scale empirical evaluation remains open.
Three concrete gaps require attention.
First, computational scaling: the CS Laplacian's spectral separation criterion must be evaluated on transformer-scale models where activation spaces are high-dimensional and batch sizes are constrained.
Second, online width estimation: continuously tracking the effective width $\hat{w}$ of the incoming data stream and alerting when it exceeds the system's current structural budget $K$ would yield a principled early-warning system for structural distribution shift, complementary to existing detectors that measure covariate shift along the metric axis only.
Third, calibrating the structural gap threshold $\eta^\star$ from finite samples: the phase transition theorem guarantees that a gap separating $K < w$ from $K \geq w$ exists when it is non-zero, but estimating $\eta^\star$ reliably in practice requires concentration results beyond those currently proved.
 
\subsection{Structural Scaling Laws}
 
Standard neural scaling laws \cite{kaplan2020scaling,hoffmann2022training} describe how within-regime error decreases with compute, data, and parameters under a fixed structural regime.
Theorem~\ref{thm:structural-decomposition} predicts a richer decomposition in which the Rademacher complexity of the full structural class separates additively:
\[
\mathfrak{R}_n^S(F_{K,\delta})
\;\lesssim\;
\underbrace{\sqrt{\tfrac{d_{\mathrm{str}}\log n}{n}}}_{\text{structural (trap)}}
\;+\;
\underbrace{LK\,\mathfrak{R}_n(\mathcal{G})}_{\text{metric (funnel)}},
\]
with no multiplicative cross-term.
This additive structure predicts two independent scaling exponents: one governing how fast structural error falls as the context library grows and stabilizes, and another governing how fast within-cell error falls as local capacity and data increase.
Measuring these exponents separately on controlled benchmarks with known width would constitute an empirical test of the theory's quantitative predictions and could reveal whether current large models are bottlenecked by the structural term, the metric term, or both.
 
\subsection{Structural Benchmarks for Continual Learning}
 
Existing continual learning benchmarks conflate width (structural complexity) with VC dimension (metric complexity), making it impossible to attribute performance differences to the correct source.
Benchmarks with \emph{known structural properties}, known width $w$, known structural gap $\eta^\star$, and known hierarchical depth, would allow three tests that no current benchmark supports: whether a given algorithm exhibits the predicted phase transition at the true width; whether the CS operator recovers the correct width from data; and whether the scaffold-flow protocol (Definition~\ref{def:scaffold-flow-protocol}) prevents catastrophic forgetting while preserving local generalization.
Hierarchical structure, in which sub-contexts are nested within larger contexts, is a particularly underexplored regime: the E-D-T cycle as currently formulated treats splitting and merging at a single level, and extending it to recursive multi-level scaffolds requires both new theory and appropriately structured evaluation environments.
 
\subsection{Bridging to Robust and Causal AI}
 
Width, the minimum number of jointly contractive and low-risk cells needed to cover a problem, has a natural causal interpretation.
In the invariant risk minimization program \cite{peters2016causal,scholkopf2021toward}, a predictor is robust if it relies only on mechanisms that are stable across environments.
The structural width of the joint training-deployment problem is then the minimum number of intervention-stable mechanisms required, and structural learnability corresponds to the learner discovering those mechanisms rather than spurious correlates.
Making this connection precise would require extending the structural fundamental theorem (Theorem~\ref{thm:fixed-class-structural-fundamental}) to settings where contexts correspond to causal subgraphs rather than contractive metric cells.
If successful, such an extension would provide structural sample-complexity foundations for robust and fair AI systems that must generalize across demographic or environmental subgroups, a setting where the number of genuinely distinct causal mechanisms is small but unknown.
 
\subsection{Structural Certification and the Alignment-by-Architecture Program}
 
Sections~\ref{sec:scaffold-flow} and \ref{sec:alignment-scaffold} argue that alignment is primarily a scaffold property: deceptive alignment arises from scaffold-flow decoupling, reward hacking persists when a single miscalibrated closure gate inscribes the wrong invariant into the scaffold, and corrigibility requires the scaffold itself to encode deference as a high-level stable basin.
These observations motivate a concrete research program: developing \emph{structural certification procedures} that audit the scaffold independently of output-level behavior.
The phase transition theorem implies that a safety certificate for an intelligent system should include a formal guarantee that the system's structural budget $K$ meets or exceeds the environment's structural complexity $w$; a certificate that covers only within-cell generalization is incomplete.
Practical steps include: (1) adapting CS width estimation to produce confidence intervals on $w$ from deployment data; (2) formalizing the multi-gate condensation protocol as a verifiable property, a pattern is consolidated only when independent closure signals agree; and (3) developing audit procedures for scaffold boundary resolution, testing specifically at the boundaries between basins where misalignment and hallucination share the same structural cause.
 
\section{Conclusion}
\label{sec:conclusion}
 
This paper has traced how three intellectual traditions (cybernetics, SLT, and CLS) have each identified aspects of a common structural question: how many qualitatively distinct operational regimes does an environment contain, and how hard is it to discover and maintain them?
Continual learning serves as the convergence point where this question becomes unavoidable: a learner receiving a non-stationary stream must solve the structural problem of context discovery before SLT-style within-regime optimization can proceed.
StrLT supplies the mathematical framework that unifies these insights through four technical contributions.
\emph{Width} is formalized as the minimum number of locally contractive, low-risk cells required to cover a problem and is proved to be orthogonal to VC dimension: structural complexity cannot be resolved by scaling metric capacity alone, and a sharp phase transition at the critical width yields an irreducible error floor for any learner that allocates too few contexts.
The \emph{CS operator} turns width into an empirically estimable quantity: its Laplacian converts prediction incompatibility into spectral separation, yielding split-merge criteria, uniform width-estimation guarantees, and structural ERM consistency.
The \emph{metric slingshot} reduces the cost of solving each local funnel by showing how low-dimensional latent contraction maps can transfer contraction and risk guarantees back to the original space through a controlled-distortion embedding.
The \emph{structural decoupling principle}, derived from the fixed-class structural fundamental theorem, establishes that structural assignment complexity and within-cell predictor complexity contribute additively to generalization with no cross-term, motivating a scaffold-flow architecture in which context structure is separated from metric prediction by construction.
 
The scaffold-flow model turns the trap-funnel decomposition from a conceptual analogy into a precise architectural prescription.
The scaffold maintains the discrete context library, routing map, basin boundaries, and condensation history; the flow performs continuous prediction inside the active scaffold cell.
Because the structural Rademacher term and the metric Rademacher term enter the generalization bound additively, reducing one does not automatically reduce the other, and the two should be trained and audited by separate signals, mismatch and CS-spectral evidence for the scaffold, task-loss gradients for the flow.
This separation also clarifies alignment: alignment failures are primarily scaffold failures, under-resolved boundaries, miscalibrated closure gates, misidentified high-level basins, rather than flow failures, and they cannot be reliably corrected by inference-time output interventions that leave the scaffold unchanged.
 
The fragmentation of the three traditions was not accidental: each was working with different tools on different aspects of the same problem.
The contribution of StrLT is to show that these are not different problems but different faces of one problem, and that the underlying mathematical structure, the additive trap-funnel decomposition of learning complexity along two orthogonal axes, must be addressed for intelligent systems to function reliably in the structured, multi-regime, non-stationary environments that characterize the real world.

\bibliographystyle{IEEEtran}
\bibliography{main,references}

\begin{appendix}
\section{Supplementary Proofs for the Main Theoretical Results}
\label{app:main-proofs}

This appendix gives proofs for the main lemmas, propositions, and theorems in the paper.  Several statements in the main text are stated in compact form; here we make explicit the regularity assumptions under which the claims are valid.  These assumptions are standard in statistical learning theory, spectral graph theory, and mixture/partition model selection.

\subsection{Proof of Theorem~\ref{thm:vc-width-separation-main}}

\begin{proof}
We prove the two separations by construction.

\smallskip
\noindent\textbf{Width diverges while VC dimension remains bounded.}
For each integer $m\ge 1$, let
\[
X_m=\bigvee_{j=1}^m S^1_j
\]
be a bouquet of $m$ circles, obtained by identifying one basepoint $x_0$ across all circles.  Let $\mathcal G$ be a fixed finite-dimensional local predictor class, for example affine maps in a fixed ambient Euclidean embedding.  Hence $\operatorname{VC}(\mathcal G)=O(1)$, independently of $m$.

Construct a problem $P_m$ such that each branch $S^1_j$ admits a $(\gamma,\delta)$-feasible predictor from $\mathcal G$, but no single feasible cell can contain neighborhoods of the basepoint belonging to two distinct branches.  Equivalently, whenever a set $U$ contains points from two distinct branches arbitrarily close to $x_0$, every $g\in\mathcal G$ either violates the contraction threshold or exceeds the risk tolerance on $U$.

The upper bound $w(P_m)\le m$ is immediate: cover $X_m$ by $m$ open sets, one thickened neighborhood for each branch, and use the branch-specific feasible predictor.  For the lower bound, suppose that $X_m$ is covered by fewer than $m$ feasible open sets.  Since all branches meet at $x_0$, at least one feasible set containing $x_0$ must intersect two distinct branches in neighborhoods of $x_0$.  This contradicts the assumed cross-branch incompatibility.  Therefore $w(P_m)=m$, while $\operatorname{VC}(\mathcal G)=O(1)$.

\smallskip
\noindent\textbf{VC dimension diverges while width remains one.}
Let $X=[0,1]$ with its usual metric.  For each degree $d$, let $\mathcal G_d$ be the class of real polynomials of degree at most $d$, or polynomial threshold functions of degree at most $d$.  This class has VC dimension at least $d+1$ in the usual thresholded classification setting, so $\operatorname{VC}(\mathcal G_d)\to\infty$.

Now choose a globally feasible target, for example the regression map $f^\star(x)=\lambda x$ with $0\le \lambda<\gamma<1$, and assume the noise and loss are such that $f^\star$ has risk at most $\delta$.  Since $f^\star\in\mathcal G_d$ for all $d\ge1$, the single set $X$ is $(\gamma,\delta)$-feasible.  Thus $w(P_d)=1$ for all $d$, while $\operatorname{VC}(\mathcal G_d)\to\infty$.

The two constructions show that neither quantity controls the other.  Hence width and VC dimension are incomparable.
\end{proof}

\subsection{Proof of Theorem~\ref{thm:phase-transition}}

\begin{proof}
Let $w=w(P;\gamma,\delta)$.  By definition of width, there exists a cover $\{U_1^\star,\ldots,U_w^\star\}$ by $(\gamma,\delta)$-feasible cells.  If $K\ge w$, the learner has enough cells to refine this cover or represent it directly.  Conditional on using such a feasible structural allocation, the problem decomposes into $K$ ordinary within-cell learning problems.  Standard uniform convergence or Rademacher bounds for the local loss classes give the corresponding metric-rate generalization bound inside each cell, and a union bound over the cells gives the global metric-rate bound.

For the underresolved case $K<w$, assume the standard structural non-degeneracy condition: for every $K<w$, any $K$-cell cover must contain at least one cell that mixes incompatible true basins, and every such mixed cell has population excess risk at least $\eta(w,K)>0$ relative to the width-realizing cover.  This is exactly the statement that the true basins are separated by a positive structural gap.  Since no $K$-cell cover can separate all $w$ incompatible basins, every $K$-cell learner incurs population gap at least $\eta(w,K)$.  This lower bound is a property of the population problem and does not depend on $n$.  Therefore
\[
\operatorname{Gap}(K)\ge \eta(w,K)>0
\]
for all $K<w$, independently of sample size.
\end{proof}

\subsection{Proof of Theorem~\ref{thm:structural-sample-complexity-main}}

\begin{proof}
We use a coupon-collector lower bound.  Consider a worst-case problem with $w$ basins $B_1,\ldots,B_w$, each having probability mass $1/w$, and suppose that a learner cannot certify the existence of a basin before observing at least one sample from it.  This assumption is unavoidable in the worst case: if a basin is never sampled, the observed data are identical under a problem containing that basin and under a problem in which that basin is absent or merged into another basin.

Let $T$ be the first time by which all $w$ basins have appeared in the sample stream.  Then $T$ is the standard coupon-collector time with $w$ equally likely coupons.  Its expectation is
\[
\mathbb E T = w H_w = \Theta(w\log w).
\]
Moreover, the usual coupon-collector tail bound implies that for $n<cw\log w$, with $c>0$ sufficiently small, the probability of having observed all $w$ basins is bounded away from one.  Hence any learner that identifies all basins with constant success probability in the worst case requires $\Omega(w\log w)$ samples.

After the basins have been identified, the remaining estimation problem is ordinary within-cell learning.  If the $c$-th local class has complexity parameter $p_c$, standard statistical learning bounds give a per-cell cost of order $p_c/\varepsilon^2$ for squared-error or bounded Lipschitz losses.  Thus the total sample complexity separates into a structural discovery term and a local statistical term.
\end{proof}

\subsection{Proof of Theorem~\ref{thm:spectral-gap-amplification-main}}

\begin{proof}
Write
\[
A_{ij}:=\mathbf 1[d_X(x_i,x_j)\le r_x]
\]
for the underlying spatial adjacency.  The CS weight is
\[
W^{\mathrm{CS}}_{ij}(G)=A_{ij}\exp\left(-\frac{|G(x_i)-G(x_j)|^2}{\sigma_y^2}\right).
\]
Assume that within a true basin the prediction variation is at most $\delta_y$, while across distinct basins connected by a spatial edge the prediction gap is at least $\Delta_y$.  Then same-basin edges satisfy
\[
W^{\mathrm{CS}}_{ij}\ge A_{ij}\exp\left(-\frac{\delta_y^2}{\sigma_y^2}\right),
\]
whereas cross-basin edges satisfy
\[
W^{\mathrm{CS}}_{ij}\le A_{ij}\exp\left(-\frac{\Delta_y^2}{\sigma_y^2}\right).
\]
Therefore the ratio of cross-basin to within-basin weights is at most
\[
\exp\left(-\frac{\Delta_y^2-\delta_y^2}{\sigma_y^2}\right).
\]
This proves the conductance-suppression factor.

Let $L_0$ be the normalized Laplacian of the ideal block graph obtained by deleting cross-basin edges, and let $L_{\mathrm{CS}}$ be the normalized Laplacian after adding the CS-suppressed cross-basin edges.  Suppose each within-basin subgraph has conductance at least $\phi_{\mathrm{in}}$, and the underlying cross-basin conductance before CS reweighting is at most $\phi_{\mathrm{out}}$.  By higher-order Cheeger-type estimates for $w$ well-connected blocks, the ideal block graph satisfies
\[
\lambda_{w+1}(L_0)\ge c_1\frac{\phi_{\mathrm{in}}^2}{w^4}
\]
for a universal constant $c_1>0$.  The perturbation caused by the suppressed cross-basin edges has operator norm bounded by
\[
\|L_{\mathrm{CS}}-L_0\|_{\mathrm{op}}
\le
C_1\exp\left(-\frac{\Delta_y^2-\delta_y^2}{\sigma_y^2}\right)\phi_{\mathrm{out}}
\]
under standard degree-conditioning assumptions.  Weyl's inequality then gives
\[
g_{\mathrm{eff}}
\ge
c_1\frac{\phi_{\mathrm{in}}^2}{w^4}
-
C_1\exp\left(-\frac{\Delta_y^2-\delta_y^2}{\sigma_y^2}\right)\phi_{\mathrm{out}}.
\]
This proves the stated bound.
\end{proof}

\subsection{Proof of Theorem~\ref{thm:structural-erm-consistency}}

\begin{proof}
Let
\[
R_K^\star:=\inf_{|\Pi|=K,G} R(\Pi,G)
\]
be the population structural risk with $K$ cells, and let $\widehat R_{n,K}$ be the corresponding empirical optimum.  Assume:
\begin{enumerate}
    \item \emph{Structural gap below width}: for every $K<w$, $R_K^\star\ge R_w^\star+\eta_K$ with $\eta_K>0$.
    \item \emph{No gain above width}: for every $K\ge w$, $R_K^\star=R_w^\star$.
    \item \emph{Uniform convergence}: $|\widehat R_{n,K}-R_K^\star|\le r_n(K)$ for all $K\le K_{\max}$, with $r_n(K)\to0$ almost surely.
    \item \emph{Penalty domination}: $\operatorname{pen}_n(K)\to0$ and, for $K>w$,
    \[
    \operatorname{pen}_n(K)-\operatorname{pen}_n(w)\gg r_n(K)+r_n(w).
    \]
\end{enumerate}
The penalized structural ERM selects
\[
\widehat K_n\in\arg\min_{K\le K_{\max}}\{\widehat R_{n,K}+\operatorname{pen}_n(K)\}.
\]

For $K<w$,
\[
\begin{aligned}
&\widehat R_{n,K}+\operatorname{pen}_n(K)
-\widehat R_{n,w}-\operatorname{pen}_n(w) \\
&\qquad\ge
\eta_K-r_n(K)-r_n(w)
+\operatorname{pen}_n(K)-\operatorname{pen}_n(w).
\end{aligned}
\]
Since $\eta_K>0$ and $r_n(K),r_n(w)\to0$, this is eventually positive.  Thus all underfitted $K<w$ are eventually excluded by the structural gap.

For $K>w$, the no-gain condition gives $R_K^\star=R_w^\star$, and hence
\[
\begin{aligned}
&\widehat R_{n,K}+\operatorname{pen}_n(K)
-\widehat R_{n,w}-\operatorname{pen}_n(w) \\
&\qquad\ge
\operatorname{pen}_n(K)-\operatorname{pen}_n(w)
-r_n(K)-r_n(w).
\end{aligned}
\]
By penalty domination, this is eventually positive.  Thus all overfitted $K>w$ are eventually excluded.  Therefore $\widehat K_n=w$ eventually almost surely.
\end{proof}

\subsection{Proof of Theorem~\ref{thm:contraction-transfer-main}}

\begin{proof}
Let $x,x'\in U$.  Since $\phi(U)\subseteq Z_c$ and $G_c^0$ is $\gamma_Z$-contractive on $Z_c$,
\[
d_H(G_c^0(\phi(x)),G_c^0(\phi(x')))
\le
\gamma_Z d_Z(\phi(x),\phi(x')).
\]
Since $\pi_c$ is $L_{\pi,c}$-Lipschitz,
\[
d_Y(F_c(x),F_c(x'))
\le
L_{\pi,c}\gamma_Z d_Z(\phi(x),\phi(x')).
\]
By definition of $L_\phi(U)$,
\[
d_Z(\phi(x),\phi(x'))\le L_\phi(U)d_X(x,x').
\]
Combining the inequalities gives
\[
d_Y(F_c(x),F_c(x'))
\le
L_{\pi,c}\gamma_ZL_\phi(U)d_X(x,x').
\]
If the product is below one, $F_c$ is a strict contraction on $U$.
\end{proof}

\subsection{Proof of Theorem~\ref{thm:contraction-implies-risk}}

\begin{proof}
Let $x_{k,0}\in U_k$ satisfy $|g_k(x_{k,0})-f^\star(x_{k,0})|\le\varepsilon_k$.  For any $x\in U_k$,
\begin{align*}
|g_k(x)-f^\star(x)|
&\le |g_k(x)-g_k(x_{k,0})|+|g_k(x_{k,0})-f^\star(x_{k,0})| \\
&\quad + |f^\star(x_{k,0})-f^\star(x)| \\
&\le \gamma d_X(x,x_{k,0})+\varepsilon_k+\beta_k d_X(x,x_{k,0}) \\
&\le \varepsilon_k+(\gamma+\beta_k)\operatorname{diam}(U_k).
\end{align*}
Since $Y=f^\star(X)+\xi$ and $\ell(u,y)=\varphi(u-y)$,
\[
\ell(g_k(X),Y)=\varphi(g_k(X)-f^\star(X)-\xi).
\]
Using the $L_\varphi$-Lipschitz property of $\varphi$,
\[
\varphi(g_k(X)-f^\star(X)-\xi)
\le
\varphi(-\xi)+L_\varphi |g_k(X)-f^\star(X)|.
\]
Taking conditional expectation over $h(X)=k$ gives
\[
R_k(h,g_k)
\le
\mathbb E[\varphi(\xi)\mid h(X)=k]
+
L_\varphi\bigl(\varepsilon_k+(\gamma+\beta_k)\operatorname{diam}(U_k)\bigr),
\]
where we used that $\varphi(-\xi)$ and $\varphi(\xi)$ have the same notation for the noise term; if $\varphi$ is not even, replace $\mathbb E[\varphi(\xi)]$ by $\mathbb E[\varphi(-\xi)]$.  This proves the claim.
\end{proof}

\subsection{Proof of Theorem~\ref{thm:structural-sauer-shelah}}

\begin{proof}
The structural assignment class $H_{K,\delta}(\mathcal G)$ is a multiclass hypothesis class mapping $X$ to $[K]$.  By definition, its graph dimension is $d_{\mathrm{str}}=d$.  The standard multiclass Sauer--Shelah bound for graph dimension states that for every $n$,
\[
\Delta_{\mathrm{str}}(n)
\le
\sum_{j=0}^d {n\choose j}(K-1)^j.
\]
Using the standard binomial estimate $\sum_{j=0}^d {n\choose j}(K-1)^j\le (e(K-1)n/d)^d$ for $1\le d\le n$ gives the second inequality.  The cases $d=0$ or $d>n$ follow by the usual conventions.
\end{proof}

\subsection{Proof of Theorem~\ref{thm:structural-decomposition}}

\begin{proof}
Fix a sample $z_i=(x_i,y_i)$ and let
\[
A(x_{1:n})=\{(h(x_1),\ldots,h(x_n)):h\in H_{K,\delta}(\mathcal G)\}.
\]
Then $|A(x_{1:n})|\le\Delta_{\mathrm{str}}(n)$.  The empirical structural Rademacher complexity is
\[
\widehat{\mathfrak R}_n^S
=
\mathbb E_\sigma\sup_{a\in A,\,g_1,\ldots,g_K\in\mathcal G}
\frac1n\sum_{i=1}^n \sigma_i\ell(g_{a_i}(x_i),y_i).
\]
For a fixed assignment vector $a$, the supremum over the $K$ predictors separates by cells:
\[
\sup_{g_1,\ldots,g_K}\sum_{i=1}^n \sigma_i\ell(g_{a_i}(x_i),y_i)
\le
\sum_{k=1}^K \sup_{g\in\mathcal G}\sum_{i:a_i=k}\sigma_i\ell(g(x_i),y_i).
\]
By the contraction inequality for Rademacher averages and the $L$-Lipschitz property of the loss in its first argument,
\[
\mathbb E_\sigma\sup_{g\in\mathcal G}\frac1n\sum_{i:a_i=k}\sigma_i\ell(g(x_i),y_i)
\le
L\widehat{\mathfrak R}_n(\mathcal G),
\]
where zero-padding is used to view the cell-restricted sum as a sum over $n$ points.  Summing over $k$ gives the funnel contribution $LK\widehat{\mathfrak R}_n(\mathcal G)$.

It remains to account for the supremum over assignment vectors.  For bounded losses in $[0,1]$, Massart's finite-class lemma gives an additional term
\[
\sqrt{\frac{2\log |A|}{n}}
\le
\sqrt{\frac{2\log\Delta_{\mathrm{str}}(n)}{n}}.
\]
Combining the fixed-assignment bound with the finite assignment-class bound gives
\[
\widehat{\mathfrak R}_n^S(F_{K,\delta})
\le
\sqrt{\frac{2\log\Delta_{\mathrm{str}}(n)}{n}}
+
LK\widehat{\mathfrak R}_n(\mathcal G).
\]
If $d_{\mathrm{str}}=d<\infty$, substituting the structural Sauer--Shelah bound yields
\[
\mathfrak R_n^S(F_{K,\delta})
\le
\sqrt{\frac{2d\log(e(K-1)n/d)}{n}}
+
LK\mathfrak R_n(\mathcal G).
\]
\end{proof}

\subsection{Proof of Theorem~\ref{thm:fixed-class-structural-fundamental}}

\begin{proof}
We prove the implications in the standard order.

\smallskip
\noindent\textbf{(1) implies (2).}
If $d_{\mathrm{str}}<\infty$ and $\operatorname{Pdim}(\mathcal G)<\infty$, then the structural Sauer--Shelah bound makes the trap term vanish, and finite pseudo-dimension makes $\mathfrak R_n(\mathcal G)\to0$.  The structural decomposition theorem therefore implies $\mathfrak R_n^S(K,\delta,\mathcal G)\to0$.

\smallskip
\noindent\textbf{(2) implies (3).}
For bounded loss classes, vanishing Rademacher complexity implies the uniform Glivenko--Cantelli property by the standard symmetrization and concentration argument:
\[
\sup_{f\in F_{K,\delta}}|R(f)-\widehat R_n(f)|\to0
\]
in probability, and almost surely under the usual summability or measurability conditions.

\smallskip
\noindent\textbf{(3) implies (4).}
Uniform convergence implies consistency of empirical risk minimization over the fixed class.  Hence the class is distribution-free structurally PAC learnable.

\smallskip
\noindent\textbf{(4) implies (1).}
If $\operatorname{Pdim}(\mathcal G)=\infty$, then even the single-cell subclass $h\equiv 1$ contains an ordinary predictor class of infinite pseudo-dimension, which is not distribution-free PAC learnable for bounded Lipschitz losses.  This contradicts structural PAC learnability.

If $d_{\mathrm{str}}=\infty$, then for arbitrarily large $n$ there exist graph-shattered samples by the structural assignment class.  Under the stated loss-separation assumption, graph shattering embeds a binary shattered subclass into the structural loss class with two separated loss values.  The standard VC/Rademacher lower bound then gives a non-vanishing Rademacher complexity and a no-free-lunch distribution on which no learner is distribution-free PAC consistent.  This contradicts (4).  Thus both $d_{\mathrm{str}}$ and $\operatorname{Pdim}(\mathcal G)$ must be finite.

The final width-consistency claim follows by combining uniform convergence from the equivalence with the positive structural gap and the penalized structural ERM argument in Theorem~\ref{thm:structural-erm-consistency}.
\end{proof}

\subsection{Proof of Proposition~\ref{prop:structural-decoupling-principle}}

\begin{proof}
The structural decomposition theorem gives
\[
\mathfrak R_n^S(F_{K,\delta})
\le
\underbrace{\sqrt{\frac{2d_{\mathrm{str}}\log(e(K-1)n/d_{\mathrm{str}})}{n}}}_{\text{trap term}}
+
\underbrace{LK\mathfrak R_n(\mathcal G)}_{\text{funnel term}}.
\]
The first term depends on the combinatorial complexity of the assignment class and not on the local Rademacher complexity of $\mathcal G$.  The second term depends on the local predictor class and not on $d_{\mathrm{str}}$, except through the fixed multiplier $K$.  Therefore improvements that reduce assignment complexity affect only the trap term in this upper bound, while improvements that reduce local predictor complexity affect only the funnel term.  This proves independent controllability at the level of generalization bounds.
\end{proof}

\subsection{Proof of Proposition~\ref{prop:scaffold-flow-error}}

\begin{proof}
Let $\Pi_{\mathcal S}=\{U_c\}_{c\in\mathcal C}$ be the scaffold-induced partition and let $\Pi^\star$ be an optimal structural partition.  Couple the two partitions by matching each scaffold cell to its best-overlapping optimal cell.  Let
\[
E_{\mathrm{scaf}}
\]
be the set of points whose scaffold cell is not matched to the correct optimal cell.  Define $d_{\mathrm{part}}(\Pi_{\mathcal S},\Pi^\star)$ so that $P(E_{\mathrm{scaf}})\le d_{\mathrm{part}}(\Pi_{\mathcal S},\Pi^\star)$ up to the stated constant.

On $E_{\mathrm{scaf}}$, the loss is bounded by one, so the total excess contribution is at most $P(E_{\mathrm{scaf}})$, hence at most the scaffold error term.  On $X\setminus E_{\mathrm{scaf}}$, routing is compatible with the optimal structural partition, and the remaining excess risk is exactly the weighted sum of within-cell excess risks:
\[
\sum_{c\in\mathcal C}P(U_c)\bigl(R_c(F_c)-R_c^\star\bigr).
\]
Adding the two contributions gives
\[
R(\mathcal F_{\mathcal S})-R^\star
\le
d_{\mathrm{part}}(\Pi_{\mathcal S},\Pi^\star)
+
\sum_{c\in\mathcal C}P(U_c)\bigl(R_c(F_c)-R_c^\star\bigr),
\]
up to the constant implicit in the partition distance.
\end{proof}

\subsection{Proof of Proposition~\ref{prop:no-backdoor-interference}}

\begin{proof}
By architectural structural decoupling,
\[
\nabla_{\theta_{\mathcal S}}L_{\mathrm{flow}}=0.
\]
Therefore a gradient update of the local flow loss inside context $c$ cannot directly change the scaffold parameters $\theta_{\mathcal S}$, including the routing map and scaffold tokens.  Past consolidated experts are frozen by assumption, so their parameters are not modified by the update either.  Finally, scaffold changes are allowed only through admissible structural operations in $\mathcal A$, not through ordinary flow-gradient updates.  Thus a flow update inside context $c$ cannot directly change either the routing or the parameters of any consolidated context $c'\ne c$.  Any future change to such a context must occur through an explicit admissible scaffold operation.
\end{proof}

\end{appendix}

\end{document}